\documentclass[
]{ceurart}
\pdfoutput=1

\usepackage{listings}

\begin{document}

\copyrightyear{2023}
\copyrightclause{Copyright for this paper by its authors.
  Use permitted under Creative Commons License Attribution 4.0
  International (CC BY 4.0).}

\conference{SeWebMeDa'23: 6th Workshop on Semantic Web solutions for large-scale biomedical data analytics}

\title{On the Potential of Artificial Intelligence Chatbots for Data Exploration of Federated Bioinformatics Knowledge Graphs}

\author[1]{Ana Claudia Sima}[%
orcid=0000-0003-3213-4495,
email=Ana-Claudia.Sima@sib.swiss
]
\author[1]{Tarcisio Mendes de Farias}[%
orcid=0000-0002-3175-5372,
email=Tarcisio.Mendes@sib.swiss
]
\address[1]{SIB Swiss Institute of Bioinformatics, Lausanne, Switzerland}

\begin{abstract}
  In this paper, we present work in progress on the role of artificial intelligence (AI) chatbots, such as ChatGPT, in facilitating data access to federated knowledge graphs. In particular, we provide examples from the field of bioinformatics, to illustrate the potential use of Conversational AI to describe datasets, as well as generate and explain (federated) queries across datasets for the benefit of domain experts.
\end{abstract}


\maketitle

\section{Introduction}

The sheer growth in number of scientific datasets published yearly shows that centralized approaches, which suffer from poor scalability, maintenance and data redundancy issues, will likely not be the solution in going forward. As an example of  this growth, a total of over 1700 databases are available in the 2023 Nucleic Acids Research Molecular Biology Database Collection \cite{rigden20232023}. According to Sever et al. \citep{sever2023we}, the future of open research data access and retrieval is federated. However, federated data access also comes with great challenges. Bringing it into daily practice for domain experts will require improvements at many layers of the technology stack, including better federated query plans and user-facing services. These will play a crucial role in abstracting away the growing complexity in data access, thus enabling larger-scale (re)use of the federated datasets. In this paper, we argue that recent advancements in large language models and Conversational Artificial Intelligence (AI) technologies, also known as AI chatbots, such as ChatGPT, can play a role in facilitating data access, even in the more challenging case of federated knowledge graphs.


We provide an insight, through selected examples from the SIB Swiss Institute of Bioinformatics\footnote{\url{https://www.sib.swiss}}, into the areas where Conversational AI can assist researchers in benefiting from the wealth of public bioinformatics data. The SIB is a federated institution by excellence, curating a growing catalog of interoperable bioinformatics knowledge graphs\footnote{See examples at \url{https://www.expasy.org/search/sparql}}. We show how an AI chatbot can be used to leverage both public documentation and human expert input for three key areas of FAIR (Findable, Accessible, Interoperable, and Reusable) data exploration: 1) summarising dataset contents in a high-level description, understandable for end users --- i.e., contributing towards data Findability; 2) explaining example queries provided as input by users; and 3) generating (federated) queries across public knowledge graphs based on natural language questions provided by users --- i.e., facilitating Accessibility and Reuse. We note that Interoperability is a pre-requisite, given that federated queries cannot be generated across non-interoperable datasets. 

Further narrowing down the scope, although AI chatbots have already shown promising results in a wide range of tasks, little attention has been given so far to the role of their recent advancements, such as ChatGPT, in question answering over knowledge graphs. To-date, only a preprint is available \cite{omar2023chatgpt}, which compares ChatGPT \textit{against} traditional Question Answering Systems. In contrast, in this paper, we provide preliminary encouraging findings indicating that a technology like ChatGPT could be employed \textit{directly} as the translator of natural language questions to knowledge graph queries, even in the more challenging case of federated queries. These federated queries have not been addressed by traditional Question Answering Systems yet. Here, we will focus on SPARQL 1.1 federated queries \citep{world2013sparql}.

We also discuss some of the current limitations of ChatGPT, including the well-known problem of \textit{hallucinations}, \textit{i.e.}, generating inaccurate, ``made up'' answers. It is clear that a purely neural language model, with (currently) no deductive reasoning power, will be prone to mistakes and therefore not the definitive solution. However, the question we aim to answer is: given the currently narrow landscape of tools leveraging federated knowledge graphs, can technologies like ChatGPT support domain experts in exploring the available data?


\section{Use Cases}
 To perform our experiments, we relied on ChatGPT. We used the March 14\textsuperscript{th}, free research version, available online at \url{https://chat.openai.com/chat}. This version is based on the GPT-4 language model.  
\subsection{Dataset Descriptions}

In this section, we introduce the three main bioinformatics datasets we explore in this study: Uniprot \cite{uniprot2019uniprot}, OMA \cite{altenhoff2021oma} and Bgee \cite{bastian2021bgee}. We chose to focus on these datasets as they were previously used in developing one of the few existing benchmarks of bioinformatics questions and corresponding federated SPARQL queries \cite{sima2019enabling}, a benchmark developed in collaboration with domain experts. All the questions and corresponding queries in this benchmark can be used for further experimentation.

As opposed to the conventional approach to presenting these datasets, and aligned with the goals of this study, we provide their description directly from the output of ChatGPT. Due to space constraints, below we remove some of the paragraphs and provide the full output as Supplementary Material \cite{ana_claudia_sima_2023_7768342}. In \textbf{Bold}, we provide the question, as it was asked to ChatGPT. 

In these first examples, we note that the descriptions generated are a clear and understandable summary of the contents of each dataset. While for the most part the information is factually accurate, we have marked in \textit{Italics} false statements generated by the model (these are first examples of \textit{hallucinations}). We provide explanations about these as footnotes. 

\begin{itemize}
    \item \textbf{What is UniProt?}

\textbf{ChatGPT:} UniProt is a comprehensive database of protein sequences and functional information. It is a collaborative effort between the European Bioinformatics Institute (EBI), the Swiss Institute of Bioinformatics (SIB), and the Protein Information Resource (PIR) in the United States. The name ``UniProt'' stands for ``Universal Protein Resource.''

UniProt is a central repository for protein sequence and annotation data from a variety of sources, including experimental data, computational predictions, and literature curation.




    \item \textbf{Does UniProt have a SPARQL endpoint?}

\textbf{ChatGPT:} Yes, UniProt provides a SPARQL endpoint that allows users to query UniProt data using the SPARQL query language. The UniProt SPARQL endpoint is available at http://sparql.uniprot.org/sparql.

The UniProt SPARQL endpoint allows users to query a wide range of UniProt data, including protein sequences, annotations, and features. Users can use SPARQL to retrieve specific data from UniProt and integrate it with data from other sources.

\textit{UniProt also provides a web interface called ``UniProt RDF Platform'' that allows users to explore and query UniProt data using SPARQL queries. The UniProt RDF Platform is available at https://www.uniprot.org/rdf/. The platform includes a query builder and several pre-defined queries that users can modify and execute.}\footnote{It is unclear why the model generates this erroneous information; however, one plausible explanation is a confusion with the European Bioinformatics Institute (EBI) RDF Platform.}

The UniProt SPARQL endpoint \textit{and RDF Platform} are important resources for researchers and bioinformaticians who need to integrate UniProt data with other biological data sources for analysis and research purposes.
    \item \textbf{What is OMA? Does it have a SPARQL endpoint?}

\textbf{ChatGPT:} OMA (Orthologous MAtrix) is a database that provides comprehensive and accurate orthology predictions across a wide range of eukaryotic organisms. OMA uses a graph-based algorithm to identify orthologous groups of genes and proteins, and it takes into account both sequence similarity and evolutionary distances between species. OMA provides a SPARQL endpoint that allows users to query the database using the SPARQL query language. The OMA SPARQL endpoint is available at http://sparql.omabrowser.org/sparql.

Users can use SPARQL to retrieve specific data from OMA, including orthologous groups, gene and protein sequences, and evolutionary distances between species. The OMA SPARQL endpoint allows users to integrate OMA data with other biological data sources and perform complex queries and analysis.


    \item \textbf{What is Bgee and does it have a SPARQL endpoint?}

\textbf{ChatGPT:} Bgee is a database that provides gene expression data for multiple species, with a focus on vertebrates. Bgee integrates and harmonizes expression data from multiple sources, including RNA sequencing and microarray experiments, and provides a user-friendly web interface for data exploration and analysis. Bgee is developed and maintained by the Swiss Institute of Bioinformatics (SIB) and is updated regularly with new data and features. \textit{Bgee currently covers over 9,000 species, including mammals, birds, reptiles, amphibians, and fish.}\footnote{This is a gross error, as the number of species in Bgee is only 52 currently. Dynamically evolving statistics are in general not reliably captured in ChatGPT.} Bgee provides a SPARQL endpoint that allows users to query the database using the SPARQL query language. The Bgee SPARQL endpoint is available at http://bgee.org/sparql.



\end{itemize}

\subsection{Explaining existing queries}
First, we note that, to our knowledge, no dedicated tools for explaining (federated) SPARQL queries are publicly available. We use ChatGPT without any fine-tuning for explaining queries, both over a single dataset (UniProt), as well as over the federation of UniProt, OMA and Bgee.

\subsubsection{Explaining queries for single dataset}

For this first exercise we chose UniProt, as it provides good documentation online, with many available examples. We prompted ChatGPT to explain a simple UniProt SPARQL query that asks for 10  example bacterial taxa in the database\footnote{A modified version of \url{https://sparql.uniprot.org/.well-known/sparql-examples/?offset=1}, restricting the list of prefixes to the strictly necessary ones (``up:'', ``taxon:'' and ``rdfs:'') and limiting the number of results to 10. Full queries and answers for the examples in this paper are provided in our Supplementary Material \cite{ana_claudia_sima_2023_7768342}}. 

\paragraph{ChatGPT UniProt Query Explanation:}

This is a SPARQL query that retrieves information about 10 taxa (a taxonomic group) and their scientific names from a knowledge graph that includes information from UniProt, a comprehensive resource for protein sequence and functional information.

Here is a breakdown of the query:

The first three lines are PREFIX declarations that define prefixes for the namespaces used in the query: up: is a prefix for the UniProt core namespace
[...]
The third line specifies that the \texttt{?taxon} must be a subclass of the taxonomic group with ID taxon:2, which \textit{represents the group of cellular organisms.}
Finally, the LIMIT clause specifies that only the first 10 results that satisfy the conditions should be returned.

Overall, this query retrieves the scientific names of the first 10 taxa that are subclasses of the \textit{cellular organisms group} in the UniProt knowledge graph.

\paragraph{Discussion:} Here, most of the explanation is factually correct and would provide helpful support for a user with limited knowledge of SPARQL or the UniProt data model. However, we also see a recurring problem in the generative model: identifiers are not accurately captured. This can potentially lead to wrong answers. In the UniProt query provided as input, \textit{taxon:2} refers to the bacterial taxa. These are indeed a subclass of \textit{cellular organisms}, but the generated explanation mentions these likely only by coincidence, since in most cases, identifiers are completely mismatched. 

An ad-hoc solution against identifier ``hallucinations'' is to simply ask for clarifications on how these identifiers were computed. The explanations usually point to the correct sources, even if the model does not actually ``do what it says'', (being a purely statistical model that does not perform reasoning). We provide an example in Section \ref{Limitations}.


%
%

 \subsubsection{Explaining federated queries}

The challenge here lies in the relatively rare occurrence of federated queries online, which implies less opportunity for the model to learn about their characteristics, as compared to single-target queries. We chose to illustrate here one of the more challenging queries that uses all 3 resources jointly: UniProt, OMA and Bgee. The query looks for human genes associated with cancer and their orthologs expressed in the rat brain. The query is available on BioQuery\footnote{\url{https://biosoda.expasy.org/build_biosodafrontend/}} interface (search for ``cancer'' in the input field). We reproduce below the full explanation:

\paragraph{ChatGPT federated query explanation:}

This is a complex SPARQL query that retrieves information about genes and proteins associated with cancer in humans and their orthologs in rats. It also retrieves information about genes expressed in the brain of rats.

Let's break it down:

The first five lines of the query define several prefixes that will be used throughout the query to abbreviate long URIs. This makes the query easier to read and write.

The SELECT statement at the end of the query specifies the variables that should be returned in the query results. In this case, the query will return two variables: \texttt{?gene} and \texttt{?orthologous\_protein2}.

The WHERE clause specifies the conditions that must be met in order for a triple pattern to be included in the query results.

The query has three nested sub-queries that are executed using SPARQL endpoints provided by different web services:

The first sub-query (inside the curly braces) searches the UniProt database for proteins in Homo sapiens that are associated with cancer. The \texttt{FILTER CONTAINS} function filters annotations that contain the word ``cancer''.

The second sub-query (also inside the curly braces) searches the OMA database for orthologous proteins to the proteins found in the first sub-query. The orthologous proteins must be in Rattus norvegicus, and must have a gene symbol that is related to the protein using the \texttt{sio:SIO\_010079} predicate.

The third sub-query (also inside the curly braces) searches the Bgee database for genes expressed in the brain of Rattus norvegicus. The genes must have an ortholog in the UniProt database that corresponds to the protein found in the second sub-query.

Finally, the whole query is wrapped in another SELECT statement that selects the variables \texttt{?gene} and \texttt{?orthologous\_protein2} from the results of the nested sub-queries.
Overall, this query retrieves information about genes and proteins associated with cancer in humans and their orthologs in rats, as well as genes expressed in the brain of rats.

\paragraph{Discussion:} The explanation is surprisingly good considering the complexity of this query (3 sources with more than 3 triple patterns each). It captures the essential information: the goal of the query in the very first paragraph (and repeated in the concluding sentence), the sources used (which ChatGPT successfully identified directly from their associated SPARQL endpoints, which are present in the query), as well as the purpose of each source in the context of the query (proteins with disease association for UniProt, orthology for OMA and expression for Bgee). The only important missed point is that the rat orthologous genes themselves must also be expressed in the brain (the explanation fails to make this connection). A correction would be: \textit{their orthologs in rats, which should furthermore be expressed in the brain}.

\subsection{SPARQL Query Generation}

\subsubsection{Queries over a Single Resource}

We start again with a relatively simple example, asking ChatGPT to construct a SPARQL query that counts all mouse proteins in UniProt.
\\

\textbf{Question:} Can you show me a SPARQL query that counts the number of mouse proteins in UniProt?

\textbf{ChatGPT:} Here is a SPARQL query that counts the number of mouse proteins in UniProt:

\begin{lstlisting}[language=SPARQL, morekeywords={SERVICE, COUNT, AS}, basicstyle=\footnotesize]
PREFIX up: <http://purl.uniprot.org/core/>
PREFIX taxon: <http://purl.uniprot.org/taxonomy/>
SELECT (COUNT(DISTINCT ?protein) as ?count) WHERE {
  ?protein a up:Protein ; up:organism taxon:10090 . }
\end{lstlisting}

Explanation, the query starts by defining two prefixes: up for the UniProt core ontology and taxon for the UniProt taxonomy ontology.
It then selects a count of distinct \texttt{?protein} URIs that are of type up:Protein and have an organism of type taxon:10090, which corresponds to the mouse taxonomic identifier in UniProt.
The count is returned as a variable \texttt{?count}.
Note that this query uses the UniProt SPARQL endpoint (https://sparql.uniprot.org/sparql), so you will need to have access to it to run the query.

\paragraph{Discussion:} The generated query is correct and directly executable at the UniProt SPARQL endpoint. It is worth noting that the taxonomic identifier of ``Mouse'' is also correct, showing that frequently used ones are also captured by the model. While this was a relatively simple query, which could be generated by traditional Question Answering Systems, the adjoining explanation is an important bonus for understanding the generated query. We note that existing Question Answering Systems over Knowledge Graphs do \textit{not}  describe the queries produced.

\subsubsection{Generating Federated Queries}
\label{Federated}

We have experimented with generating several federated queries based on natural language questions. Although typically they are not directly executable, these first attempts are also not very far from the ground truth. We argue that a systematic evaluation (part of future work) should take into account the number of correct triple patterns in the queries generated, rather than the number of correct queries generated. It should also consider the number of user-provided feedback steps required to arrive at the correct query. We recall here that the problem of Question Answering over Federated Knowledge Graphs has, to-date, not been tackled in the published literature.

We provide the reverse example of the previous section - namely, we ask ChatGPT to generate a federated SPARQL query using UniProt, OMA and Bgee for retrieving the rat orthologs expressed in the brain of human genes associated with cancer.  

We show the generated query below:

\begin{lstlisting}[language=SPARQL, morekeywords={SERVICE}, basicstyle=\footnotesize]
PREFIX up:<http://purl.uniprot.org/core/>
PREFIX oma:<http://omabrowser.org/ontology#>
PREFIX bgee:<http://bgee.org/ontology/bgee_ontology#>

SELECT DISTINCT ?human_gene ?rat_gene ?bgee_expr
WHERE {
  # Get human genes associated with cancer from UniProt
  SERVICE <https://sparql.uniprot.org/sparql> {
    SELECT DISTINCT ?human_gene
    WHERE {
      ?human_gene up:annotation [ up:disease [ a up:Disease ;
            up:category up:"Cancer-related genes" ] ] .}}
  
  # Get the corresponding rat genes from OMA
  SERVICE <http://sparql.orthology.stanford.edu/sparql> {
    SELECT DISTINCT ?human_gene ?rat_gene
    WHERE {
      ?human_oma_ortho oma:hasDbXref 
      up:HGNC_OrthologousGene [ oma:hasDbXref ?human_gene ] ;
        oma:hasDbXref up:Rat_OrthologousGene 
        [ oma:hasDbXref ?rat_gene ] .}}
  
  # Get the expression levels of the rat genes in the brain from Bgee
  SERVICE <http://bgee.org/sparql> {
    SELECT DISTINCT ?rat_gene ?bgee_expr
    WHERE {
      ?rat_gene bgee:organ [ bgee:organName "brain" ] ;
                bgee:expression [ bgee:expressionLevel ?bgee_expr ] .}}}
\end{lstlisting}

\paragraph{Discussion:} Perhaps unsurprisingly, the UniProt part of the query is almost correct, the only modification required being that ``Cancer-related genes'' is too specific and a filter on the annotation would be better than an exact match. Interestingly, the OMA SPARQL endpoint generated in this example is erroneous and points to a non-existing URL. The query is also incorrect. For Bgee, while the SPARQL endpoint URL is correct, the query fragment is not. 

As a step to improve the query, we supplied the correct OMA query fragment manually in a few different formulations. We note that this is a particularly complex fragment, as OMA structures orthologous genes as Hierarchical Orthologous Groups, which need to be traversed through SPARQL property paths. We asked ChatGPT to use this in order to rephrase the original response. Surprisingly, this feedback only resulted in an improvement when the variable names themselves were meaningful, showing that the model is influenced by subtle changes.

We show below the fragment that lead to an improvement of the federated query:

\begin{lstlisting}[language=SPARQL, morekeywords={SERVICE}, basicstyle=\footnotesize]

<https://sparql.omabrowser.org/sparql/> {
SELECT ?protein_human_uniprot ?protein_rat ?gene_rat WHERE {
	?protein_human a orth:Protein .
	?orthologous_protein2 a orth:Protein .
	?cluster a orth:OrthologsCluster .
	?cluster orth:hasHomologousMember ?node1 .
	?cluster orth:hasHomologousMember ?node2 .
	?node2 orth:hasHomologousMember* ?protein_human .
	?node1 orth:hasHomologousMember* ?protein_rat. 
        ?protein_rat orth:organism/obo:RO_0002162/
            up:scientificName 'Rattus norvegicus' .
	?protein_rat sio:SIO_010079 ?gene_rat .
	?protein_human lscr:xrefUniprot ?protein_human_uniprot .
	FILTER(?node1 != ?node2)  }}
\end{lstlisting}

We note that simply changing the variable names to less meaningful ones, e.g. \texttt{?protein\_X} and \texttt{?protein\_Z} lead to wrong results in the federated query generated by the model, pointing to issues in the stability of the results. The full interactions (feedback provided to ChatGPT and the updated answers returned) are described in our Supplementary Material \cite{ana_claudia_sima_2023_7768342}. These interactions also illustrate non-determinism in the generation of answers in response to the same question. We further discuss limitations in the following section.

\section{Limitations and Mitigations}
\label{Limitations}
Conversational AI has recently achieved impressive results, gaining attention across many disciplines \cite{oh2023chatgpt, kung2023performance}. However, it is still a recent technology, with important limitations. Here, we summarise a few of these limitations, with a focus on the use of ChatGPT for question answering over federated knowledge graphs:
\begin{itemize}
    \item Hallucinations. These are erroneous statements produced by the model and are particularly prevalent in the case of identifiers, as a statistical model cannot accurately capture these given their relatively rare occurence. A mitigation in this case is to ask a follow-up question on how the identifier was computed, as these explanations usually point to the correct sources. An example is reproduced in the following:

\textbf{Question}: What is the taxonomic ID of Colocasia esculenta?

\textbf{ChatGPT}: \textit{The taxonomic ID of Colocasia esculenta is 4472.} (note: the answer is wrong.)

\textbf{Question}: How did you compute this?

\textbf{ChatGPT}: \textit{I retrieved the taxonomic ID of Colocasia esculenta from the National Center for Biotechnology Information (NCBI) Taxonomy database. The NCBI Taxonomy database is a comprehensive resource that provides information on the classification and phylogeny of organisms. To find the taxonomic ID of Colocasia esculenta, I searched for the scientific name of the species in the database and retrieved its taxonomic information.} 

This explanation points to the correct source, and searching for \textit{Colocasia esculenta} in NCBI\footnote{\url{https://www.ncbi.nlm.nih.gov/taxonomy}} will indeed retrieve the correct ID. A future solution could be tighter integration with dedicated tools that can compute this information, via plugins.\footnote{See \url{https://openai.com/blog/chatgpt-plugins}} For example, a Python plugin, as well as a Wolfram Alpha extension, have already been implemented.

    \item Non-determinism. Different calls to the language model might result in different answers, which can be confusing for users. Interestingly, when we searched a second time for ``what is bgee?'' with less context (i.e., without prior asking about UniProt and OMA), the answer generated was less accurate (\textit{e.g.}, that Bgee stands for ``Brain Gene Expression''). 
    \item Licensing. Perhaps one of the most significant limitations is that although the research version of ChatGPT is still free, the API isn't, making reproducibility difficult. Recently launched alternatives such as the LLaMa language models from META AI \cite{touvron2023llama}, might provide a viable solution, with a free API. 
    \item Debugging answers. It is very difficult today to understand how a particular answer was generated. As we have seen with the case of hallucinations, the explanations about how a piece of information was constructed (e.g., identifiers) is not consistent with the result. This is a severe limitation for scientific research use cases.

    \item Subtle changes in the input can lead to significant changes in the generated answers. This observation is general for deep learning models and here, we have illustrated this with the example of the feedback provided to the federated query generated in Section \ref{Federated}.
    
    \item Performance of generated queries. Given that the model does not have any knowledge regarding query planning, the queries it will generate might be highly suboptimal in terms of runtime performance. Since SPARQL 1.1. federated query performance is still highly dependent on the formulation of the query, this can lead to \textit{e.g.} server timeouts.
    
\end{itemize}

%

\section{Conclusions}

Federated open research data access is likely to be the way of the future, given the continuous growth in number of datasets produced in recent years. However, bringing federated data closer to the users still requires technological improvements, including improved user-facing services. 

While we provided here only a brief overview of potential uses for Conversational AI in exploring public datasets, particularly federated knowledge graphs, the following preliminary conclusions can already be drawn:
\begin{enumerate}
    \item The importance of publicly available documentation for existing datasets is perhaps higher than ever, as it now enables easier Findability and Reuse for researchers interested in discovering public datasets on a given topic, by asking questions in natural language (e.g., to ChatGPT). Moreover, ChatGPT can provide users high-level summaries of the contents of public datasets, making it easier for researchers to understand at a glance if a given resource is relevant for their analyses. Here, we have illustrated this with the examples of the UniProt, OMA and Bgee databases. 
    \item Domain experts can use ChatGPT for explaining example SPARQL queries. Furthermore, SPARQL experts can  directly contribute towards improving the language models with feedback on queries generated by the models based on natural language descriptions, even if they do not have specific machine learning expertise. Over time, this will benefit domain experts in generating correct queries for their purposes.
    \item Despite these advantages, caution should still be exercised when using AI chatbots such as ChatGPT for accessing data, in particular due to \textit{hallucinations} - seemingly confident, but \textit{incorrect} answers to user inputs. More research should be dedicated in filtering out these cases, perhaps based on the confidence levels in the underlying language model. For now, asking the model for explanations seems to be a viable  solution for these cases. However, the responsibility of validating answers still lies mainly on the end user today.
\end{enumerate}


\bibliography{sample-ceur}


\end{document}